# Mobile Sensing for Multipurpose Applications in Transportation.

**Armstrong Aboah**
PhD Student
Civil and Environmental Engineering Department
University of Missouri-Columbia
Email: aa5mv@umsystem.edu

**Michael Boeding**
Student
Computer Science Department
University of Missouri-Columbia
Email: michaelboeding@mail.missouri.edu

**Yaw Adu-Gyamfi**
Assistant Professor
Civil and Environmental Engineering Department
University of Missouri-Columbia
Email: adugyamfiy@missouri.edu

## ABSTRACT
Routine and consistent data collection is required to address contemporary transportation issues. The cost of data collection increases significantly when sophisticated machines are used to collect data. Due to this constraint, State Departments of Transportation struggle to collect consistent data for analyzing and resolving transportation problems in a timely manner. Recent advancements in the sensors integrated into smartphones have resulted in a more affordable method of data collection. The primary objective of this study is to develop and implement a smartphone application for data collection. The currently designed app consists of three major modules: a frontend graphical user interface (GUI), a sensor module, and a backend module. While the frontend user interface enables interaction with the app, the sensor modules collect relevant data such as video and accelerometer readings while the app is in use. The backend, on the other hand, is made up of firebase storage, which is used to store the gathered data. In comparison to other developed apps for collecting pavement information, this current app is not overly reliant on the internet enabling the app to be used in areas of restricted internet access. The developed application was evaluated by collecting data on the i-70W highway connecting Columbia, Missouri, and Kansas City, Missouri. The data was analyzed for a variety of purposes, including calculating the International Roughness Index (IRI), identifying pavement distresses, and understanding driver's behaviour and environment . The results of the application indicate that the data collected by the app is of high quality.

## INTRODUCTION
To address contemporary transportation issues, massive amounts of data must be collected. The cost of collecting these data is enormous, and as such, less costly methods of data collection is eminent. Among these issues are analyzing pavement conditions and understanding drivers' behavior and environments. For example, road surface data must be collected and analyzed on a routine basis to ensure proper road maintenance. The International Roughness Index (IRI) is one of the metrics derived from this data . The IRI indexes the road surface's roughness. This provides information to road maintenance engineers about which roads require immediate attention and which ones do not. Collecting information about the pavement's surface using sophisticated machines such as the ARAN van is prohibitively expensive. This complicates the task of collecting road surface data on a consistent basis by the various state Departments of Transportation (DOT). For instance, the Virginia Department of Transportation (VDOT) spends $1.8 million per year collecting information about the pavement surface using high-end machines *(1)*. Likewise, highly sophisticated machines are used to collect data for pavement distress analysis. However, advancements in consumer-level technologies such as smartphones have made it possible to collect road surface data at a low cost for the purpose of analyzing pavement roughness and distress.

Smartphones contain sensors such as an accelerometer, which detects vibrations caused by a moving vehicle. These vibrations are used as a surrogate to estimate the roughness of the road. Numerous studies have been conducted to estimate the IRI using smartphone data, and the results are comparable to those obtained using high-end machines with an acceptable margin of error *(1)(2)(3)(4)*.

Additionally, understanding the environment around vehicles is a necessary requirement in achieving a safer driving experience. Advancements in smartphone camera technology have resulted in the production of high-resolution images and videos, resulting in a more affordable method of collecting naturalistic driving data. With the aid of computer vision and deep learning techniques these smartphone videos can be processed to gain a better understanding of the driver's behavior and surroundings.

The study's aim was to develop and implement a mobile data collection application for collecting information regarding road surfaces, naturalistic driving events etc. The prototype was developed using the following techniques: specification of the hardware platform, design of the software interface, design of the software sub-components, and development of the backend service. The app was created to work with iOS devices and to directly measure the speed of moving vehicles. In comparison to previous smartphone data collection apps, this current app is not overly reliant on the internet. This means that data can be gathered and stored temporarily in the app's library before being uploaded to the cloud server for storage. This feature enables data collection from roadways in remote areas where a stable internet connection is not available. Finally, the developed app was evaluated on Missouri's Interstate 70 to evaluate the accuracy of data collected by the app.

The rest of the paper is divided into the following sections. The second section reviews relevant literature. Section three contains information about the development process, including the design approach, key components, and modulus. The fourth section discusses the data collection process as it relates to the developed mobile application. Section five summarizes the quantitative findings from the collection of field data. Finally, Section six summarizes the research, results the findings, and makes recommendations for future research in section seven.

**RELATED STUDIES**
This section discusses studies that examine the use of smartphone data to solve transportation-related problems. Each study is reviewed for its purpose, data collection technique, and methodology.

The technology to accurately assess pavement roughness with inexpensive sensors has improved greatly. A probe-based monitoring system for slippery and rough road surfaces was developed by MDOT in 2010 *(5)*. The vehicle data was collected and transmitted to a backend server running on a Droid platform.It was attached to the windshield in the same manner as a navigation device. Various sensors in the vehicle and on the phone were used to collect data, including the phone's three-axis accelerometer, the external road surface temperature and humidity, and the vehicle's Controller Area Network . During a two-year period, the system was installed in two vehicles driven by MDOT personnel. Over thirty thousand miles and more than 13 gigabytes of data were captured. To represent the pavement's surface irregularities, the vertical accelerometer signal's variation was employed. The accelerometer's sample rate is 100 Hz. Data collection was followed by calibration of the accelerometer readings using a PASER system curve fitting algorithm to the Pavement Surface Evaluation and Rating (PASER) scale. Future iterations of the curve fitting algorithm may incorporate data from the MDOT's annual PASER rating study.

Researchers at Auburn University examined the use of vehicle-mounted sensors to determine the condition of the pavement *(6)*. The study's primary objective was to compute the IRI through the use of automotive sensors. The IRI was calculated using information gathered from a variety of vehicle sensors, including suspension deflection meters, accelerometers, and gyroscopes. On a 1.7-mile (2,750-meter) long test track, the National Center for Asphalt Technology conducted controlled speed tests (NCAT). The total number of vibrations in a particular section can be determined by calculating the Root Mean Square (RMS) of a signal measurement (i.e., vertical acceleration, gyroscopes, or suspension deflection). Acceleration includes a section on RMS (Root Mean Squared) (Section 3.2). Following that, the aggregated vibrations were compared to the pavement segment's authentic IRI. The root mean square of vertical accelerations was found to be the most accurate representation of the authentic IRI. It followed the same general trend as the well-known IRI, with the exception of a few expected magnitude changes. In summary, this study established that the most practical method for calculating the IRI is to use a root mean square algorithm on vertical acceleration readings.

In a pilot study, Flintsch et al. *(7)* quantified road ride quality and roughness using probing vehicles. Once again, vertical acceleration data was used to estimate vehicle vibration. At the Virginia Smart Road facility in Blacksburg, Virginia, a smoothness profile was created using an inertial-based laser profiler, and vertical accelerometer measurements were taken using an instrumented car. The frequency of the accelerometer was chosen to be 10 hertz. Additionally, GPS coordinates were collected. Acceleration data was collected during four runs on the test track. The study discovered that acceleration runs are highly reproducible. The smoothness profile and acceleration measurements are highly correlated, as determined by the coherence function analysis.

Eleonora et al. *(8)* used smartphone data to detect and analyze risky driving behaviors. The study examined critical risk indicators such as the number of aggressive driving incidents and cell phone use while driving. The study gathered data on vehicle speed, distance traveled, accelerations, turning maneuvers, braking events, and cell phone use. The study's findings indicate that distraction caused by smartphone use has a significant effect on the number of severe events occurring per kilometer and, consequently, on the relative crash risk. Additionally, smartphone sensor data can be used to accurately detect mobile phone use while driving in more than 70% of cases.

**DESIGN APPROACH**
The current app was designed to collect data from the onboard IOS sensors and video from the onboard camera and then transfer the data to a server. The current design is made up of three major modules: a frontend user interface module, a sensor module, and a backend module. Figure 1 shows the interactions and information flow between these modules.

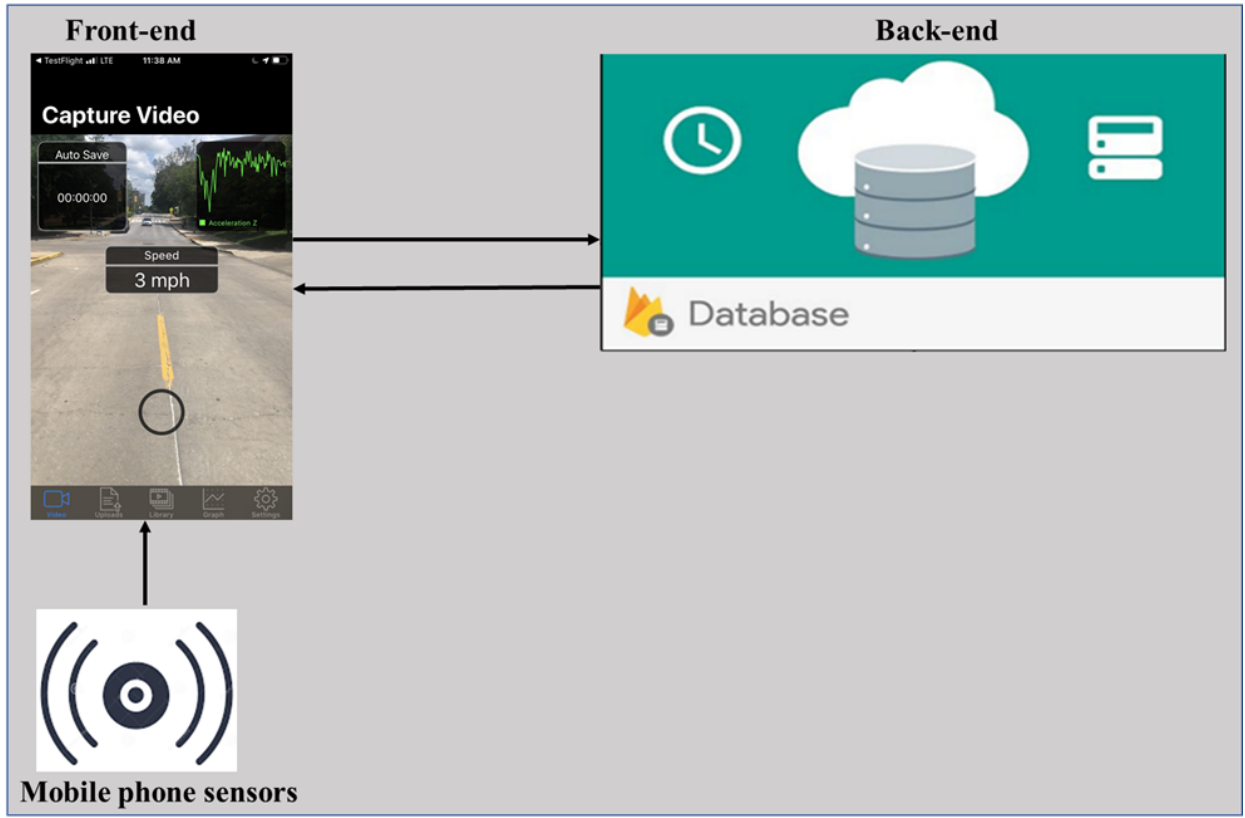

Figure 1: Frontend, and backend modules of the App

**Frontend Design**
The user interface for this app was designed to provide a streamlined view that exposed key aspects of the application via a tab bar controller. This enables us to compartmentalize the application's various critical components into distinct views. As illustrated in Figure 2, the tab bar contains a Video Tab, an Uploads Tab, a Library Tab, a Graph Tab, and a Settings Tab. The Video Tab enables the collection of video and sensor data. The Uploads Tab enables the user to keep track of the status of each package gathered in the Video Tab. The Library Tab displays all of the packages that have been collected and are currently stored on the device. The Graph Tab displays a live graph of the device's acceleration in the z direction as measured by its sensors. The Settings Tab displays all of the current settings and enables them to be modified as necessary.

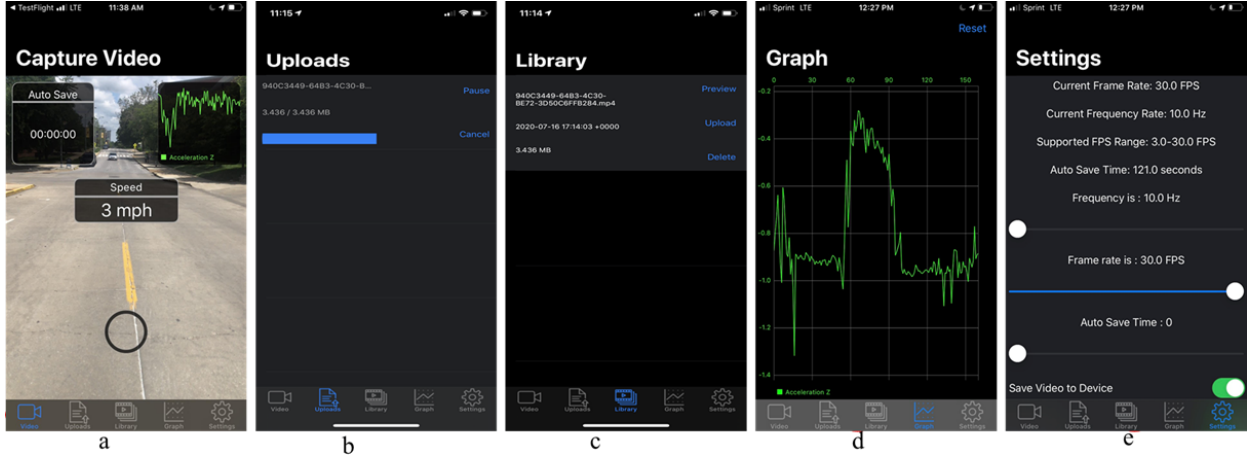

Figure 2: Key components of the app: (a) Video Tab, (b) Upload Tab, (c) Library Tab, (d) Graph Tab (e) Settings Tab

**Backend Design**
The primary function of the backend is to provide computational resources that can be used to accelerate front-end user query responses. The analytics performed on the front end of the application can be computationally expensive. To enable such sophisticated analytics on the front end of the app, we built a scalable, cloud-based backend using cutting-edge big data analytics techniques. The current study's backend is powered by Google Firebase. Firebase is made up of several parts, including a Realtime Database, a Cloud Firestore, and Cloud Storage. The Firebase Realtime Database is a relational database management system that runs in the cloud. The data is stored in JSON format and is synchronized with each connected client in real time. Similarly, the Cloud Firestore, maintains data consistency across client applications and enables offline support for mobile and web applications. Cloud Storage enables massive scalability of file storage. It allows users to upload and download files directly to and from the Cloud Storage "bucket". The developed application stores all sensor data in a Firebase Real-time Database and the videos in Cloud Firestore. The use of both the Firebase Realtime Database and Cloud Firestore ensure that uploaded data are made instantly accessible to other app users. Additionally, they allow multiple users to simultaneously push data to the cloud storage in real-time. Finally, the Cloud Firestore caches data that your app is actively using, allowing the app to write, read, listen to, and query data even when the device is not connected to the internet. The structure of the backend real-time database is shown in Figure 3a.

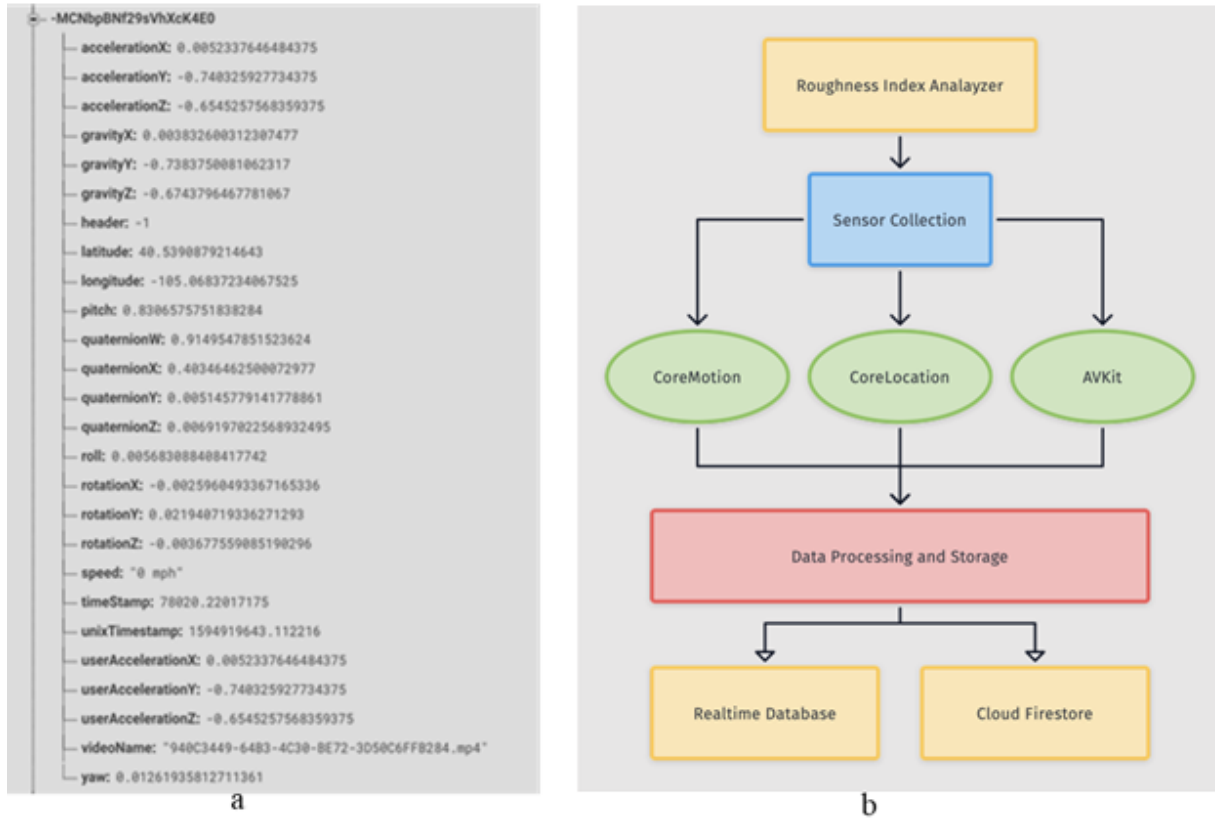

**Figure 3: (a) Real-time posting of data collected, (b) Sensor Module**

**Sensor Module**
The developed app leverages multiple modules to allow for sensor collection. These modules include CoreMotion , CoreLocation, and AVKit as shown in Figure 3b. The CoreMotion module collects motion and environmental data from the onboard hardware of iOS devices, including accelerometers and gyroscopes, as well as the pedometer, magnetometer, and barometer. This framework enables the access and utilization of data generated by the hardware. The CoreLocation determines the altitude, orientation, and geographical location of a device, as well as its position in relation to a nearby iBeacon device. The framework collects data by utilizing all available hardware components on the device, including barometer, GPS, Wi-Fi, magnetometer, and Bluetooth. Also, the AVKit provides a high-level interface for video content playback. These various modules were utilized in this current app to access the device motion data (accelerometers and gyroscopes), the GPS location, and the video data respectively.

**DATA COLLECTION USING THE ROUGHNESS INDEX ANALYZER APP**
The test for this study was conducted on I-70 West, which connects Columbia, Missouri, and Kansas City, Missouri. The road segment considered for this test was approximately 45 miles in length. The study area depicted in Figure 4 is the location of the data collection.

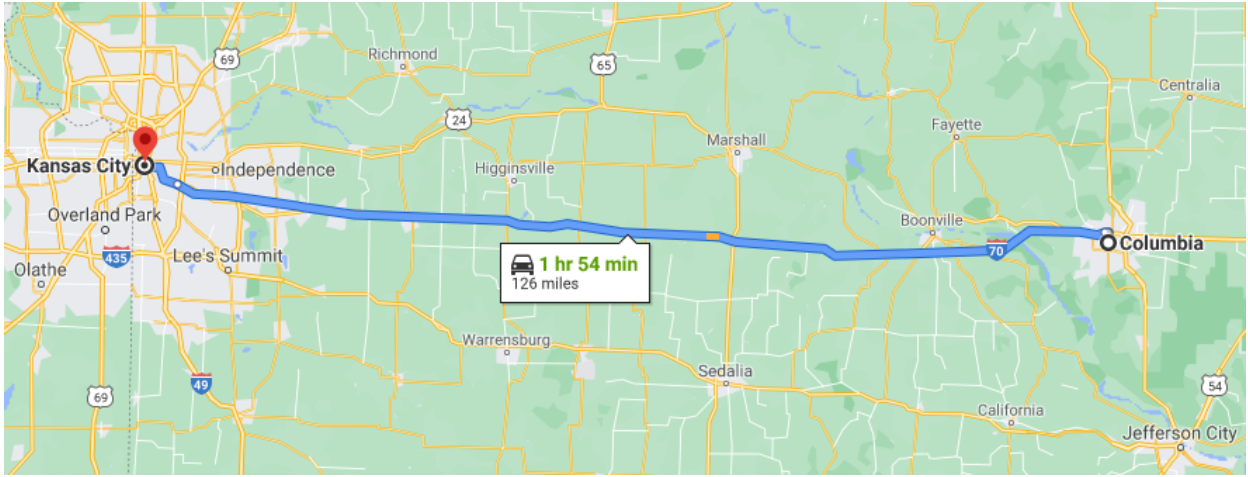

Figure 4. The study area where the app was used to collect pavement roughness information (I-70 W)

To collect data with the app, the mobile phone must first be mounted on the car's windscreen (see Figure 5b) to record the pavement surface and some other pertinent information. The video data was sampled at a frame rate of 10 frames per second, whereas the accelerometer and vehicle location data were collected at a frame rate of 30 frames per second. As illustrated in Figure 5a, the vehicle used to collect data was a 2007 Nissan Sentra.

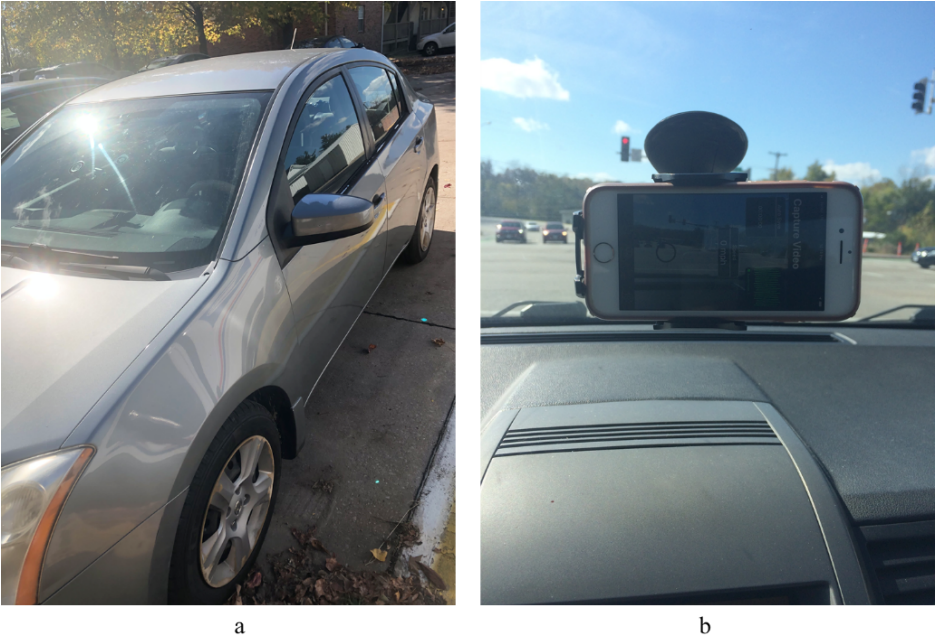

a        b

Figure 5a and 5b. a) 2007 Nissan Sentra vehicle used for the data collection b) The mounting position of the smartphone on the windscreen

**An Interactive User Interface for Data Querying using Streamlit**
The ability of the app to sync the numerous pieces of information it collects is critical. In order to achieve this, the study developed an interactive user interface for querying collected data using streamlit. Streamlit is a free and open-source Python library for web application development. Streamlit is compatible with a wide variety of popular libraries and frameworks, including Keras, OpenCV, Vega-Lite, Pytorch, Tensorflow, and Python. Due to this library's ease of use and rapid deployment, the study used it to create an interactive dashboard that enabled effortless data querying by timestamp.

The dashboard has three major menus: the homepage, the data querying page, and the data visualization page. The data querying page enables users to query both raw data and specific video frames by timestamp. On the data visualization page, users can view changes in accelerometer readings while the vehicle is in motion as illustrated in Figure 6. Acceleration changes were minimal as the vehicle came to a stop at the traffic light. Accelerometer readings shown in Figure 6 indicate that when the vehicle changes lanes, a spike is observed in the x and y directions but not in the z. However, if the vehicle passes over a pothole, spikes are seen to occur in all directions (x,y and z). This helps in differentiating between a vehicle making a lane change or negotiating a curve and a pavement irregularity.

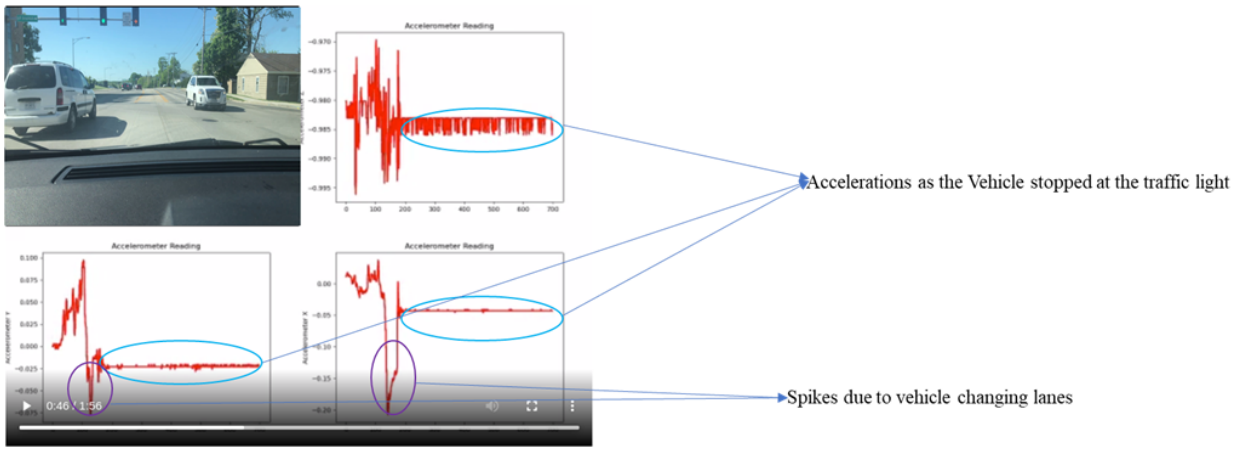

Figure 6: Changes in Accelerometer readings while vehicle in in motion

**Assessing the Accuracy of the App's GPS coordinate Points**
The GPS coordinates must be precise in order to synchronize the video and accelerometer data from the app. This will allow us to map pavement distresses in real time at the exact locations where they occurred. Issues arise when GPS information is not accurate. Figure 7 shows a Google Earth view of GPS coordinates. The GPS coordinates are observed to be slightly off, which may affect the synchronization of the collected data.

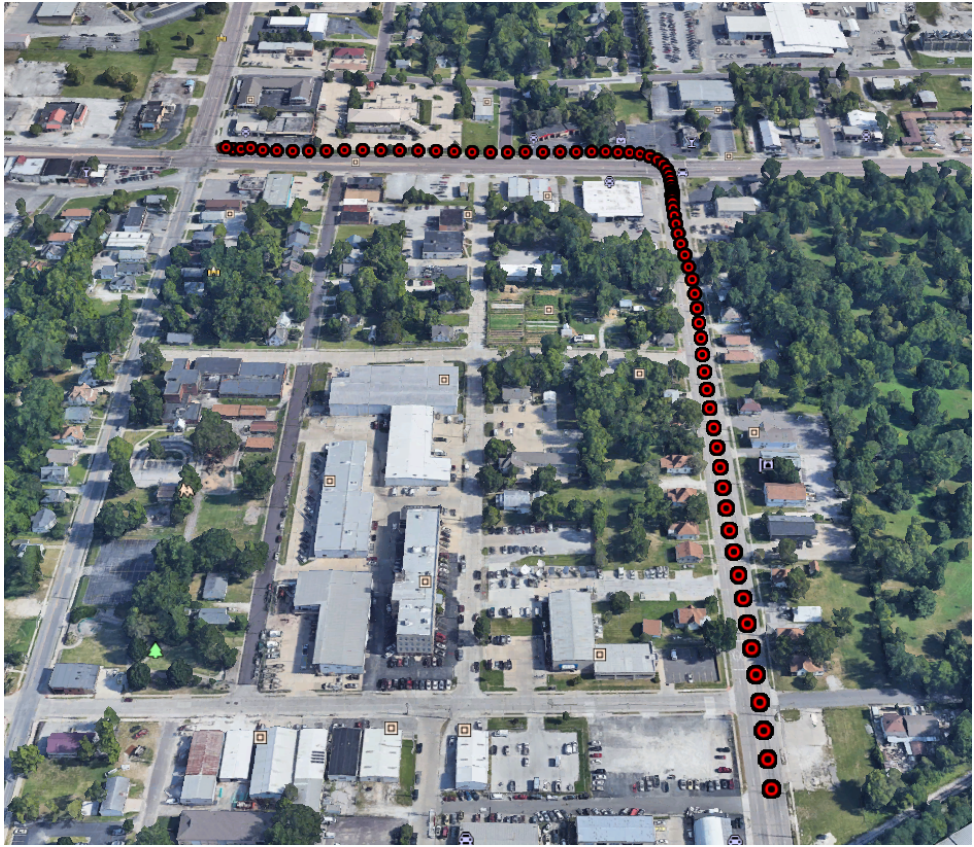

Figure 7: Plot of GPS coordinate points provided by the app

**APPLICATIONS OF DATA COLLECTED FROM THE APP**
The app collects a variety of data, including accelerometer and gyroscope readings, roadway video, and GPS location data. This section discusses some of the possible uses for the app's data.

**Estimating road roughness index (IRI)**
The accelerometer, gyroscope, and speed data collected by the app was used in a study by Aboah and Adu-Gyamfi *(2)* to forecast IRI values for road segments. In their study, the collected data was fed into a deep learning model developed to forecast the IRI value of road segments. The ground truth IRI values obtained from MoDOT's ARAN viewer portal served as the model's dependent variable *(9)*. Figure 8 depicts the ARAN viewer portal. The portal includes three tabs for pavement information: Condition, IRI, and Rut. Each tab contains records spanning the years 2009 to 2019. The portal includes a search box for locating roadways with relevant pavement information. To select a section of a roadway, the information for the road section's Start log (Begin log) and End log is entered in the search boxes shown in Figure 8. The independent variables, on the other hand, were accelerometer data, gyroscope data , and vehicle speed data from the App. As a result, the performance of the model was used to estimate the data accuracy collected by the App. The developed model's average root mean squared error (ARMSE) was 5.6. This means that our predictions are 5.6 units off the mark when compared to the actual IRI estimates provided by road profilers. Furthermore, the root-mean-squared-error (RMSPE) metric indicates that the model's predictions were approximately 17% off target IRI values. As illustrated in Figure 9a, the road conditions' trends and amplitudes are correctly matched. As illustrated in Figure 9b, there is a linear relationship between the true and predicted IRI values. The r-squared value of the linear plot was 0.79. This indicates that the true and predicted IRI values are highly correlated. Additionally, the model's r-squared value was 0.79.

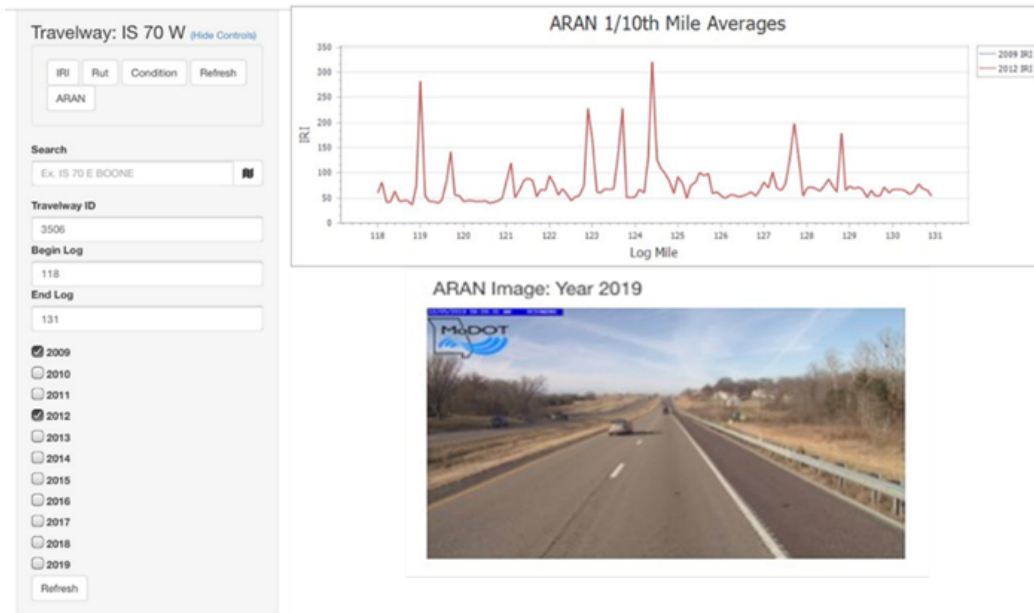

Figure 8. The MoDOT ARAN viewer portal *(8)*

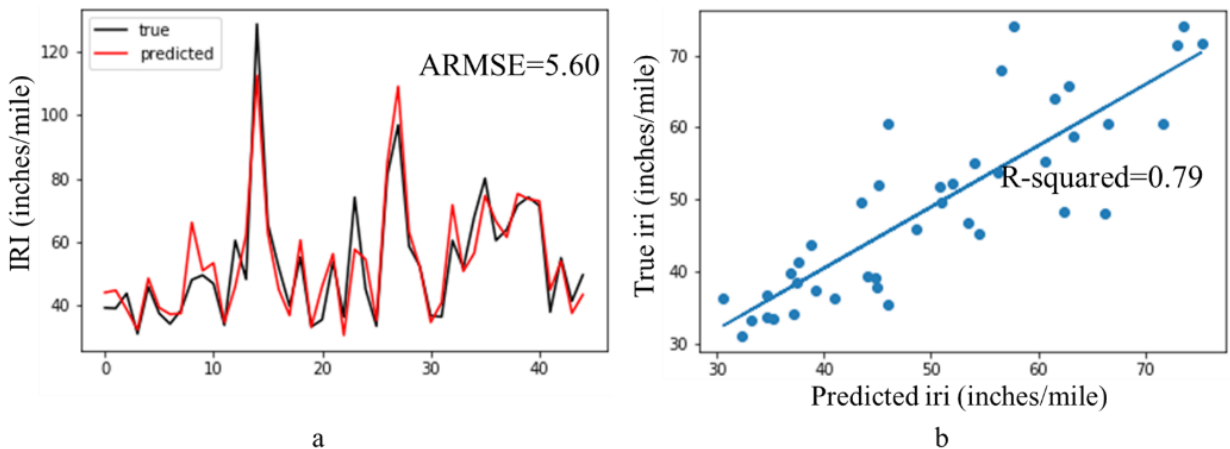

Fig. 9a and 9b. Proposed Model Performance: a). A Plot of True IRI values and Predicted IRI values and b). Scatter Plot of True IRI values against Predicted IRI values

**Pavement Distress Detection**
Pavement distresses pose a potential threat to the safety of road users. As a result, detecting distresses in a timely manner is regarded as one of the most important steps in limiting further degradation of pavement surfaces. To make the best use of financial resources, it is necessary to assess the condition of pavement surfaces on a regular basis and keep up with maintenance. Using a proposed detection model by Mandal et al. *(10)*, video data obtained from the app was used to detect pavement distress. Mandal et al. *(10)* developed a simplified system for evaluating the surface of pavements through the use of deep learning algorithms. The researchers learned the visual and textual patterns associated with the various types of distress using three single-stage object detection algorithms: YOLO, CenterNet, and EfficientDet. The developed models were evaluated for their predictive and categorizing abilities, and the F1 score was calculated using statistical precision and recall values. On two test datasets released by the IEEE Global Road Damage Detection Challenge, the top-performing model achieved F1 scores of 0.58 and 0.57. Figure 10 illustrates identified pavement distresses from video data. Three pavement distresses were identified in this figure. These were designated as D00 (red bounding box), D20 (blue bounding box), and D40 (green bounding box), respectively, to represent longitudinal crack, alligator crack, and pothole.

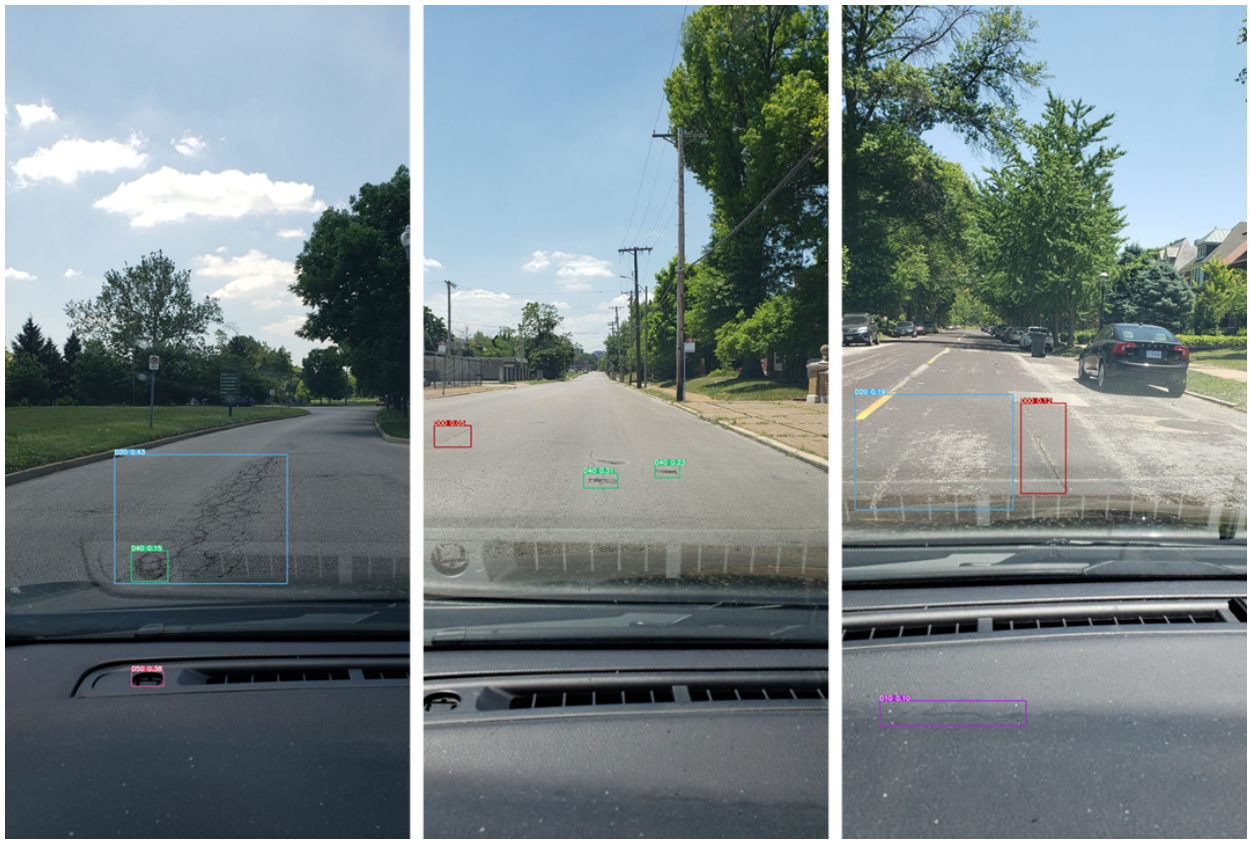

Figure 10: Pavement. distresses detected from video data

**Naturalistic Driving Studies**
Naturalistic driving studies (NDS) are cutting-edge research techniques that involve continuously recording driving data in real-world driving conditions using advanced instrumentation. NDSs enable the assessment of driving risks that would otherwise be impossible to assess using traditional crash databases or experimental methods *(11)*. The NDS findings have a significant impact on policy making, safety research and development of safety countermeasures. The developed application in this study makes use of two cameras to collect NDS data. One camera monitors the driver's environment outside the vehicle, while the other is used to record the driver's activities inside the vehicle as illustrated in Figure 11a and 11b. The collected data was analyzed to build an LSTM model for identifying various driving maneuvers as part of the process of understanding the driver's environment. The developed model was evaluated using accuracy and precision. The base model which used the sensor data and image data had an accuracy of 94% and precision of 86%. To evaluate the influence of using image information, the study constructed a second model that used an accelerometer and gyroscope sensor data. The second model had an accuracy of 90% and a precision of 83%. The model's confusion matrices are depicted in Figure 12. The model's output indicates that data collected by the app is of high-quality.

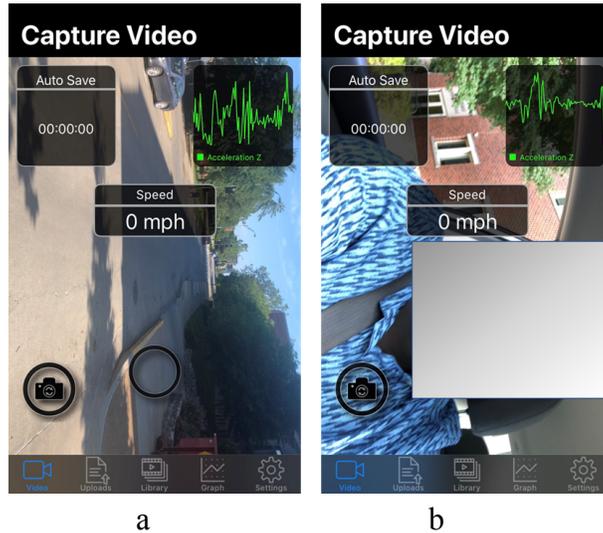

Figure 11: a)Back camera recording activities outside the vehicle b) Front camera recording activities inside the vehicle

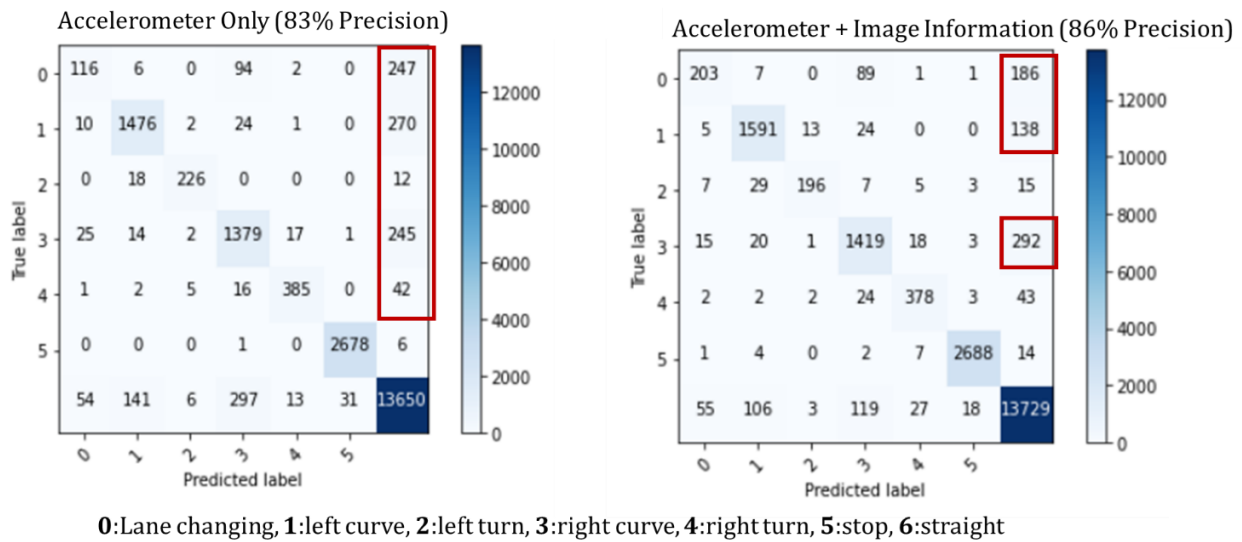

Figure 12: Confusion Matrices

## CONCLUSION

This study developed and deployed a workable mobile application for collecting data on road surface conditions. The app collects data about the road surface via in-built smartphone sensors. The data is initially stored in the app's library and is then uploaded to the cloud storage. The app's settings allow users to change the sampling frequency and frame rate during data collection. Additionally, while driving, the app is able to display the accelerometer readings on the video tab interface. The road roughness application was developed using a modular approach. The design framework of this app is composed of three major modules: a frontend user interface (UI) component, a sensor component, and a backend component.

The app's functionality was evaluated by collecting data on the road surface on the I-70 W highway connecting Columbia, Missouri and Kansas City. The collected data was used to build a predictive model for estimating IRI values using a deep learning architecture. The accuracy of the model was used as a proxy for the quality of data collected by the app. ARMSE, R-squared, and RMSPE were used as metrics for evaluation. The model predictions indicate that the predicted IRI values from the smartphone data are comparable to those estimated using high-end machines such as the ARAN van. When the predicted IRI values were compared to the ground truth IRI values, a goodness-of-fit value of 0.79 was obtained. This demonstrates a high degree of correlation between them. Also, the video information from the app can be used to identify pavement distresses.

In a nutshell, when tested, the developed mobile application for collecting road surface information and estimating road surface roughness demonstrated a high degree of potential for producing accurate and reliable results.

**LIMITATIONS AND RECOMMENDATIONS**
The developed roughness app's future updates should address the following limitations.

- The first constraint is that the app was designed exclusively for the iOS operating system. Given the large number of Android users, the app's next update should include support for the Android operating system.
- Additionally, when used, the app does not provide IRI values directly. Future updates should incorporate the developed deep learning model into the backend, allowing the app to directly predict the IRI value of road sections in real-time while in use.
- Another limitation observed is that the GPS coordinates are slightly off as shown in Figure 7. Next update of the app will have an improved GPS.

**AUTHOR CONTRIBUTION STATEMENT**
Each of the three authors made equal contribution to: conducting experiments, analysis of the experimental data, and writing the manuscript.